\documentclass[letterpaper, 10 pt, journal, twoside]{IEEEtran}

\usepackage{array,multirow,graphics}
\usepackage{graphicx}
\usepackage{times}
\usepackage{amsmath}
\usepackage{amssymb}
\usepackage{amsbsy}
\usepackage[font=footnotesize]{caption}
\usepackage{float}
\usepackage{balance}
\usepackage{url}
\usepackage[ruled,linesnumbered,longend]{algorithm2e}
\usepackage{xcolor}
\usepackage{latexsym}
\usepackage{subcaption}
\usepackage{textcomp}
\usepackage{pdfrender}
\usepackage{tabularx}
\usepackage{booktabs}
\usepackage{fancyhdr}
\let\oldmaketitle\maketitle
 
\renewcommand{\maketitle}{%
  \oldmaketitle
  \thispagestyle{fancy}
}

\begin{document}
%
\title{An Effective Multi-Cue Positioning System for Agricultural Robotics}
%
%
%

\author{Marco Imperoli$^*$, Ciro Potena$^*$, Daniele Nardi, Giorgio Grisetti and Alberto Pretto
\thanks{Manuscript received: February, 24, 2018; Revised April, 20, 2018; Accepted June, 19, 2018. This paper was recommended for publication by Editor C. Stachniss upon evaluation of the Associate Editor and Reviewers' comments.} 
\thanks{This work was supported by the EC under Grant H2020-ICT-644227-Flourish. The Authors are with the Department of Computer, Control, and Management Engineering 
``Antonio Ruberti``, Sapienza University of Rome, Italy. Email: { \{imperoli, potena, nardi, grisetti, pretto\}@diag.uniroma1.it}.}
\thanks{$^*$~These two authors contribute equally to the work}
\thanks{Digital Object Identifier (DOI): see top of this page.}
}

\maketitle

\begin{abstract}
 The self-localization capability is a crucial component for Unmanned Ground Vehicles (UGV) in farming applications. Approaches based solely on visual cues or on low-cost GPS are easily prone to fail in such scenarios. 
  In this paper, we present a robust and accurate 3D global pose estimation framework, designed to take full advantage of heterogeneous sensory data. By modeling the pose estimation problem as a pose graph optimization, our approach simultaneously mitigates the cumulative drift introduced by motion estimation systems (wheel odometry, visual odometry, \dots), and the noise introduced by raw GPS readings. Along with a suitable motion model, our system also integrates two additional types of constraints: (i) a Digital Elevation Model and (ii) a Markov Random Field assumption. 
  We demonstrate how using these additional cues substantially reduces the error along the altitude axis and, moreover, how this benefit spreads to the other components of the state. 
  We report exhaustive experiments combining several sensor setups, showing accuracy improvements ranging from 37\% to 76\% with respect to the exclusive use of a GPS sensor. We show that our approach provides accurate results even if the GPS unexpectedly changes positioning mode. The code of our system along with the acquired datasets are released with this paper.
\end{abstract}

\begin{IEEEkeywords}
Robotics in Agriculture and Forestry, Localization and Sensor Fusion
\end{IEEEkeywords}

%
\IEEEpeerreviewmaketitle

\section*{Supplementary Material}

The datasets and the project's code are available at:
\begin{quote}
  \begin{small}
    { \texttt{http://www.dis.uniroma1.it/{\texttildelow}labrococo/fsd}}
  \end{small}
\end{quote}

\section{Introduction}

\IEEEPARstart{I}{t} is commonly believed that the exploitation of autonomous robots in agriculture represents one of the applications with the greatest impact on food security, sustainability, reduction of chemical treatments, and minimization of the human effort. In this context, an accurate global pose estimation system is an essential component for an effective farming robot in order to successfully accomplish several tasks: (i) navigation and path planning; (ii) autonomous ground intervention; (iii) acquisition of relevant semantic information. However, self-localization inside an agricultural environment is a complex task: the scene is rather homogeneous, visually repetitive and often poor of distinguishable reference points. For this reason, conventional landmark based localization approaches can easily fail. Currently, most systems rely on high-end Real-Time Kinematic Global Positioning Systems (RTK-GPSs) to localize the UGV on the field with high accuracy \cite{weiss2011plant, norremark2008development}. Unfortunately, such sensors are typically expensive and, moreover, they require at least one nearby geo-localized ground station to work properly. On the other hand, consumer-grade GPSs\footnote{In this paper, we use GPS as a synonym of the more general acronym GNSS (Global Navigation Satellite System) since almost all GNSSs use at least the GPS system, included the two GNSSs used in our experiments.} usually provide noisy data, thus not guaranteeing enough accuracy and reliability for safe and effective operations. Moreover, a GPS cannot provide the full state estimation of the vehicle, i.e. its attitude, that is an essential information to perform a full 3D reconstruction of the environment.
\begin{figure}[t!]
\includegraphics[width=.41\columnwidth]{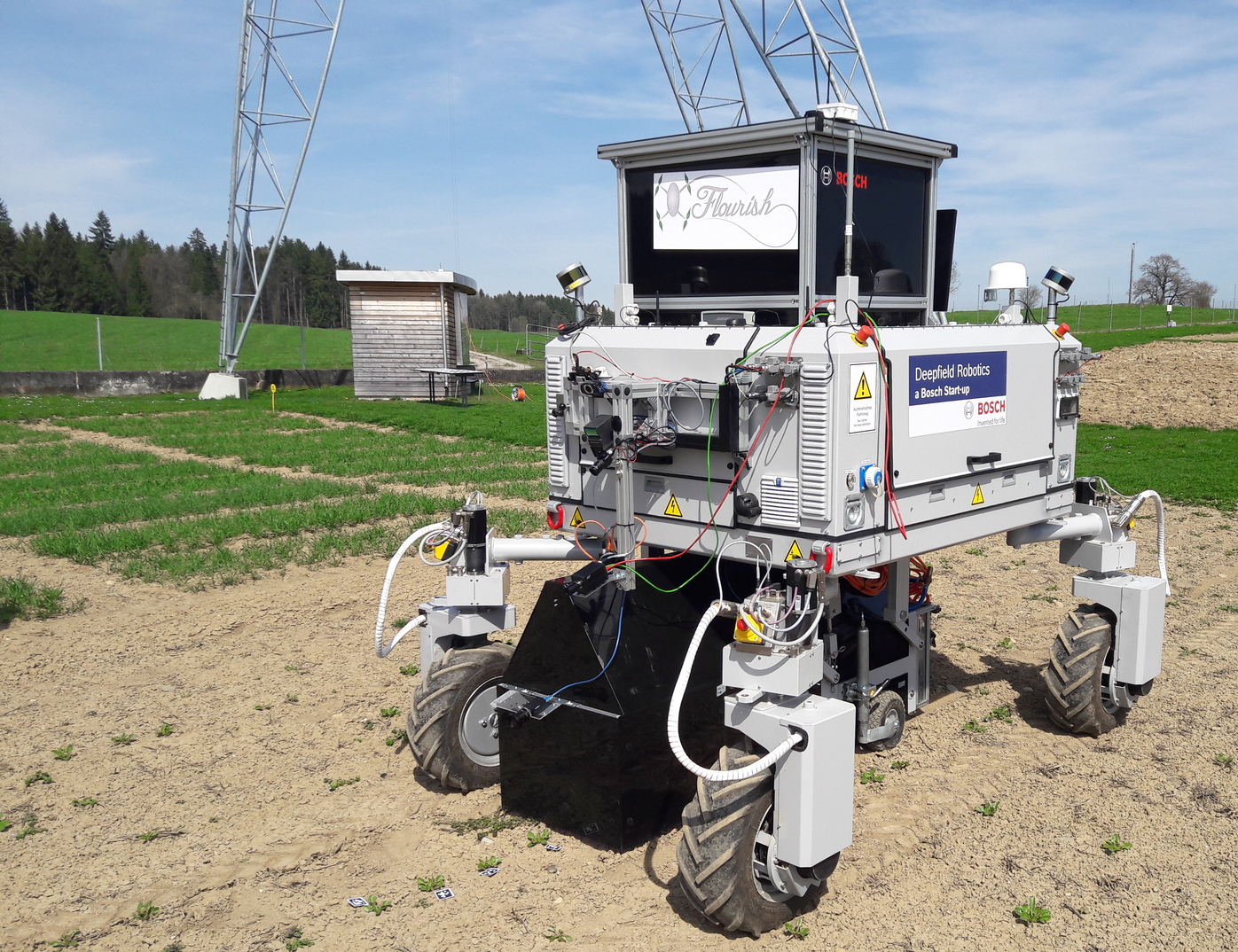} 
\includegraphics[width=.36\columnwidth]{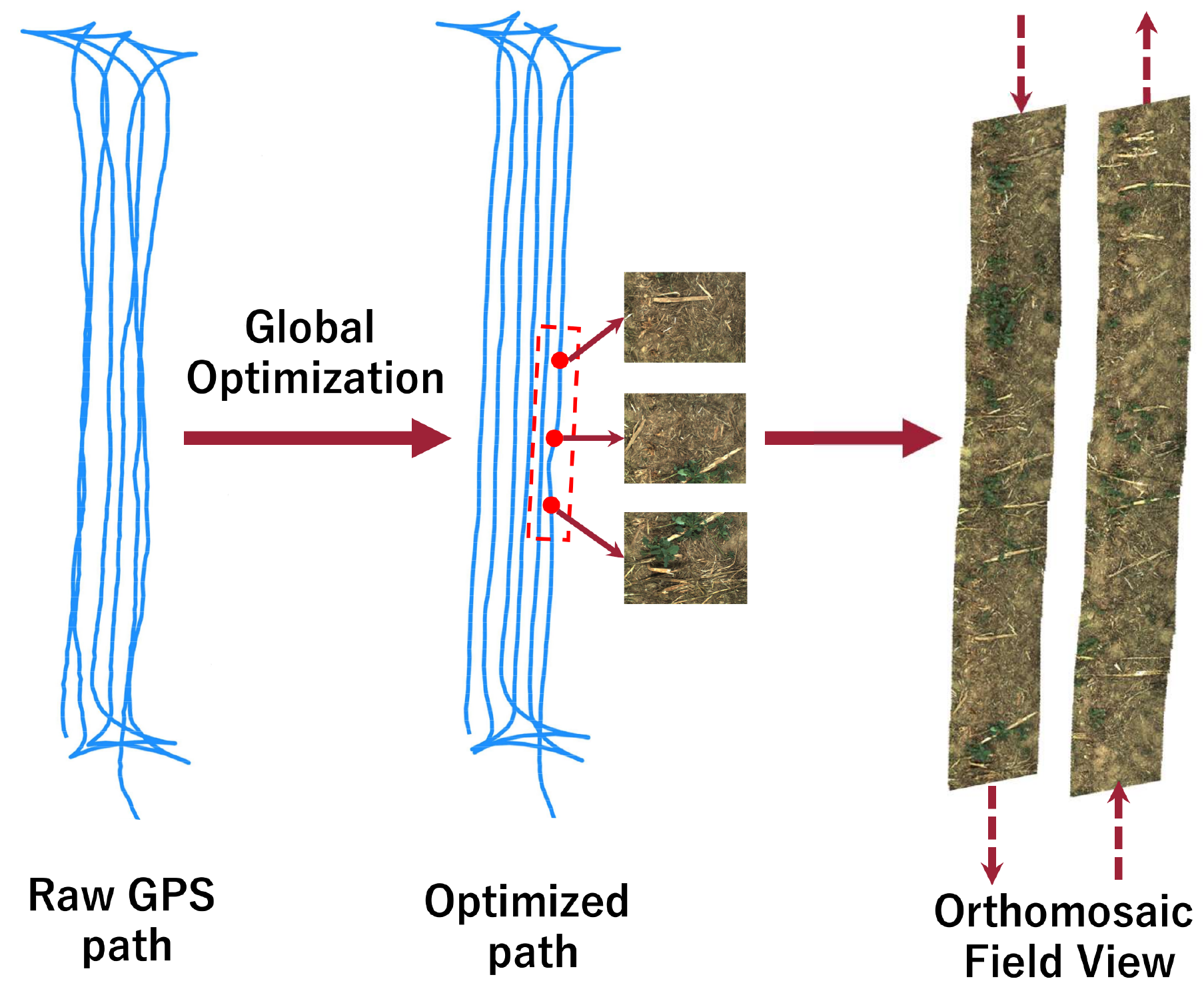}
\includegraphics[width=.315\columnwidth, angle =90]{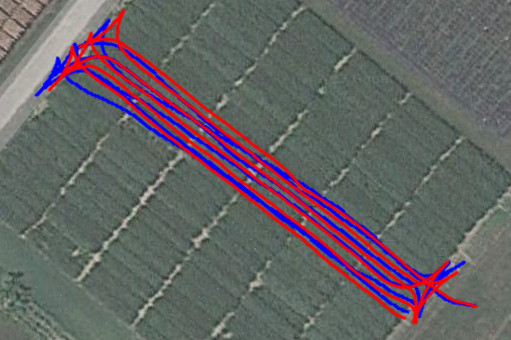} 
\caption{(Left) The Bosch BoniRob farm robot used in the experiments; 
         (Center) Example of a trajectory (Dataset B, see Sec.~\ref{sec:experiments}) optimized by using our system: the optimized pose graph can be then used, for example, to stitch together the images acquired from a downward looking camera;
         (Right) The obtained trajectory (red solid line) with respect to the trajectory obtained using only the raw GPS readings (blue solid line). Both trajectories have been over-imposed on the actual field used during the acquisition campaign.}
\label{fig:teaser}
\end{figure}
In this paper, we present a robust and accurate 3D global pose estimation system for UGVs (Unmanned Ground Vehicles) designed to address the specific challenges of an agricultural environment. Our system effectively fuses several heterogeneous cues extracted from low-cost, consumer grade sensors, by leveraging the strengths of each sensor and the specific characteristics of the agricultural context. 
We cast the global localization problem as a pose graph optimization problem (Sec.~\ref{sec:pose_graph}): the constraints between consecutive nodes are represented by motion estimations provided by the UGV wheel odometry, local point-cloud registration, and a visual odometry (VO) front-end that provides a full 6D ego-motion estimation with a small cumulative drift\footnote{In VO open-loop systems, the cumulative drift is unavoidable.}. Noisy, but drift-free GPS readings (i.e., the GPS \emph{pose solution}), along with a pitch and roll estimation extracted by using a MEMS Inertial Measurement Units (IMU), are directly integrated as prior nodes. Driven by the fact that both GPS and visual odometry provide poor estimates along the $z$-axis, i.e. the axis parallel to the gravity vector, we propose to improve the state estimation by introducing two additional altitude constraints:
\begin{enumerate}
 \item An altitude prior, provided by a Digital Elevation Model (DEM);
 \item A smoothness constraint for the altitude of adjacent nodes\footnote{The term ''adjacent`` denotes nodes that are temporally or spatially close.}.
\end{enumerate}
Both the newly introduced constraints are justified by the assumption that, in an agricultural field, the altitude varies slowly, i.e. the soil terrain can be approximated by piece-wise smooth surfaces. The smoothness constraints exploit the fact that a farming robot traverses the field by following the crop rows, hence, by using the Markov assumption, the built pose graph can be arranged as a Markov Random Field (MRF). The motion of the UGV is finally constrained using an Ackermann motion model extended to the non-planar motion case. The integration of such constraints not only improves the accuracy of the altitude estimation, but it also positively affects the estimate of the remaining state components, i.e. $x$ and $y$  (see Sec.~\ref{sec:experiments}).\\
The optimization problem (Sec.~\ref{sec:graph_optimization}) is then iteratively solved by exploiting a graph based optimization framework \cite{kummerle2011} in a sliding-window (SW) fashion (Sec.~\ref{sec:sliding_window}), i.e., optimizing the sub-graphs associated to the most recent sensor readings. The SW optimization allows to obtain on-line localization results that approximate the results achievable by an off-line optimization over the whole dataset.\\
In order to validate our approach (Sec.~\ref{sec:experiments}), we used and made publicly available with this paper two novel challenging datasets acquired using a Bosch BoniRob UGV (Fig.~\ref{fig:teaser}, left) equipped with, among several others calibrated sensors, two types of low-cost GNSSs: a Precise Point Positioning (PPP) GPS and a consumer-grade RTK-GPS. 
We report exhaustive experiments with several sensors setups, showing remarkable results: the global localization accuracy has been improved up to 37\% and 76\%, compared with the raw localization obtained by using only the raw RTK-GPS and PPP-GPS readings, respectively (e.g., Fig.~\ref{fig:teaser}). We also show that our approach allows localizing the UGV even though the GPS performances temporarily degrade, e.g. due to a signal loss.

\subsection{Related Work}

The problem of global pose estimation for UGVs has been intensively investigated, especially in the context of self driving vehicles and outdoor autonomous robots moving in urban environments. The task is commonly approached by integrating multiple sources of information. Most of the state-of-the-art systems rely on IMU-aided GPS \cite{Farrell2008}, while they differ in the other sensor cues they use in the estimation process.  Cameras are used primarily in \cite{Schleicher2009,Parra2011,Rehder–2012–7469,VishalCVPRW2015,Schreiber2016}, while LIDARs have been used in \cite{kuemmerle13icra}.\\
In urban scenarios, the presence of a prior map allows to improve the estimation by constraining the robot motion. \cite{FOUQ09aCI,Brubaker2013CVPR} use 2D road maps, while \cite{Drevelle2011-GPSSol} propose to use more rich DEMs. The sensors fusion is usually carried out by means of parametric \cite{FOUQ09aCI} or discrete \cite{kuemmerle13icra} filtering, pose graph optimization \cite{Rehder–2012–7469, VishalCVPRW2015}, set-membership positioning \cite{Drevelle2011-GPSSol}, or hybrid topological/filtering \cite{Schleicher2009}.\\ As stated in the introduction, these approaches cannot be used effectively in agricultural environments, since a prior map is typically not available. In addition, crops exhibit substantially a less stable structure than an urban environment, and their appearance varies substantially over time. 
Hence, the localization inside an agricultural field, by using a map built on-line, turns out to be extremely difficult since stable features are hard to find. For this reason, most of the available localization methods for farming robots are based on expensive global navigation satellite systems \cite{stoll2000guidance,thuilot2001automatic,norremark2008development}. However, relying on the GPS as the
primary localization sensor exposes the system to GPS related issues: potential signal losses, multi-path, and a time-dependent accuracy influenced by the satellite positions.\\
The main task of an agricultural robot is to follow the crop rows and take some action along the way.  To this extent, English~\emph{et. al}~\cite{6907079}, proposed a vision based crop-row following system. While effective, this system assumes that the crops are clearly visible from the camera of the robot, and this is not true at all growth stages of the plants.  Furthermore, the estimate of a crop row tracking tends to accumulate drift along the row direction.\\
To gain robustness and relax the accuracy requirements on the GPS, it is natural to use the plants as landmarks to build a map using a SLAM algorithm.  To this extent, Cheein \emph{et al.} \cite{Cheein2011} propose to find and to use as landmarks, in a SLAM system, olive tree stems. The stem detection algorithm uses both camera and laser data. Other approaches are based on the detection of specific plant species and thus they address very specific use cases.  Jin \emph{et al.} \cite{Jian2009} focus on the individual detection of corn plants by using RGB-D data. In \cite{weiss2011plant}, the authors propose a MEMS based 3D LIDAR sensor to map an agricultural environment by means of a per-plant detection algorithm.  Gai \emph{et al.} [6] proposed an algorithm that follows leaf ridges detected in RGB images to the center.  Similar approaches rely on Stem Emerging Points (SEPs) localizations: Mitdiby \emph{et al.} \cite{Midtiby2012} follow sugar beet leaf contours to find the SEPs.  In \cite{Haug2014} the authors perform machine learning-based SEP localization in an organic carrot field. Kraemer \emph{et al.} \cite{Kraemer2017} proposed an image-based plant localization method that exploits a CNN to learn time-invariant SEPs.

\subsection{Contributions}

In this paper, we provide a robust and effective positioning framework targeted for agricultural applications that aims to achieve high level accuracy with low cost GPSs. We integrate in an efficient way, a wide range of heterogeneous sensors into a pose graph by adapting the features of each of them to the specificity of the farming scenario. We exploit domain-specific patterns to introduce further constraints such as a MRF assumption and a DEM that contribute to the improvement of the state estimation. We evaluate our system with extensive experiments that highlight the contribution of each employed cue. We also provide an open-source implementation of our code and two challenging datasets with ground truth, acquired with a Bosch BoniRob farm robot.

\section{Multi-Cue Pose Graph}\label{sec:pose_graph}

The challenges that must be addressed to design a robust and accurate global pose estimation framework for farming applications are twofold: (i) the agricultural environment appearance, being usually homogeneous and visually repetitive; (ii) The high number of cues that have to be fused together. In this section we describe how we formulate a robust pose estimation procedure able to face both these issues.

\begin{figure}[ht]
\centering
\begin{subfigure}{0.45\columnwidth}￼
\includegraphics[width=\linewidth]{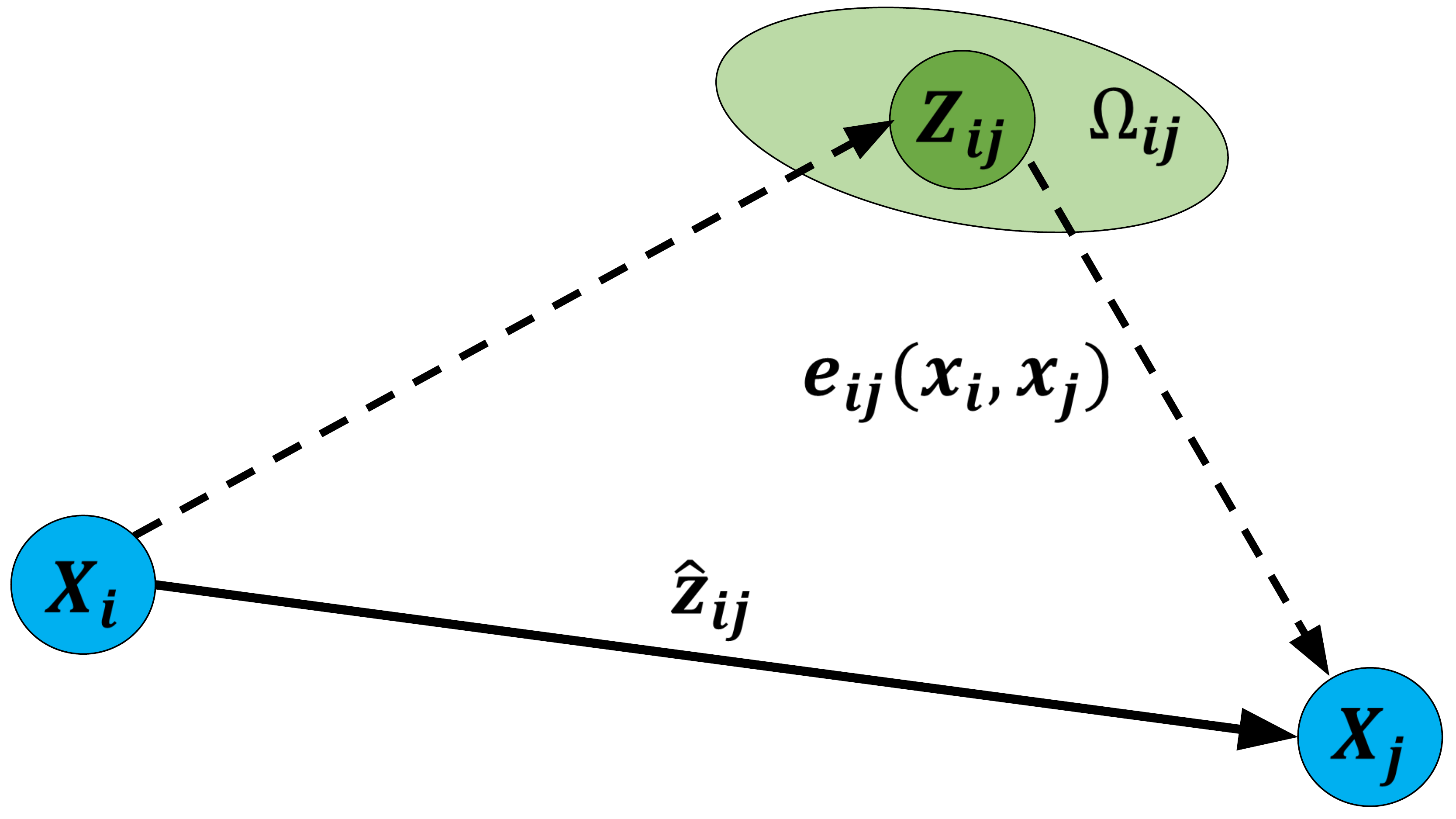}
\end{subfigure}
\begin{subfigure}{0.49\columnwidth}￼
\includegraphics[width=\linewidth]{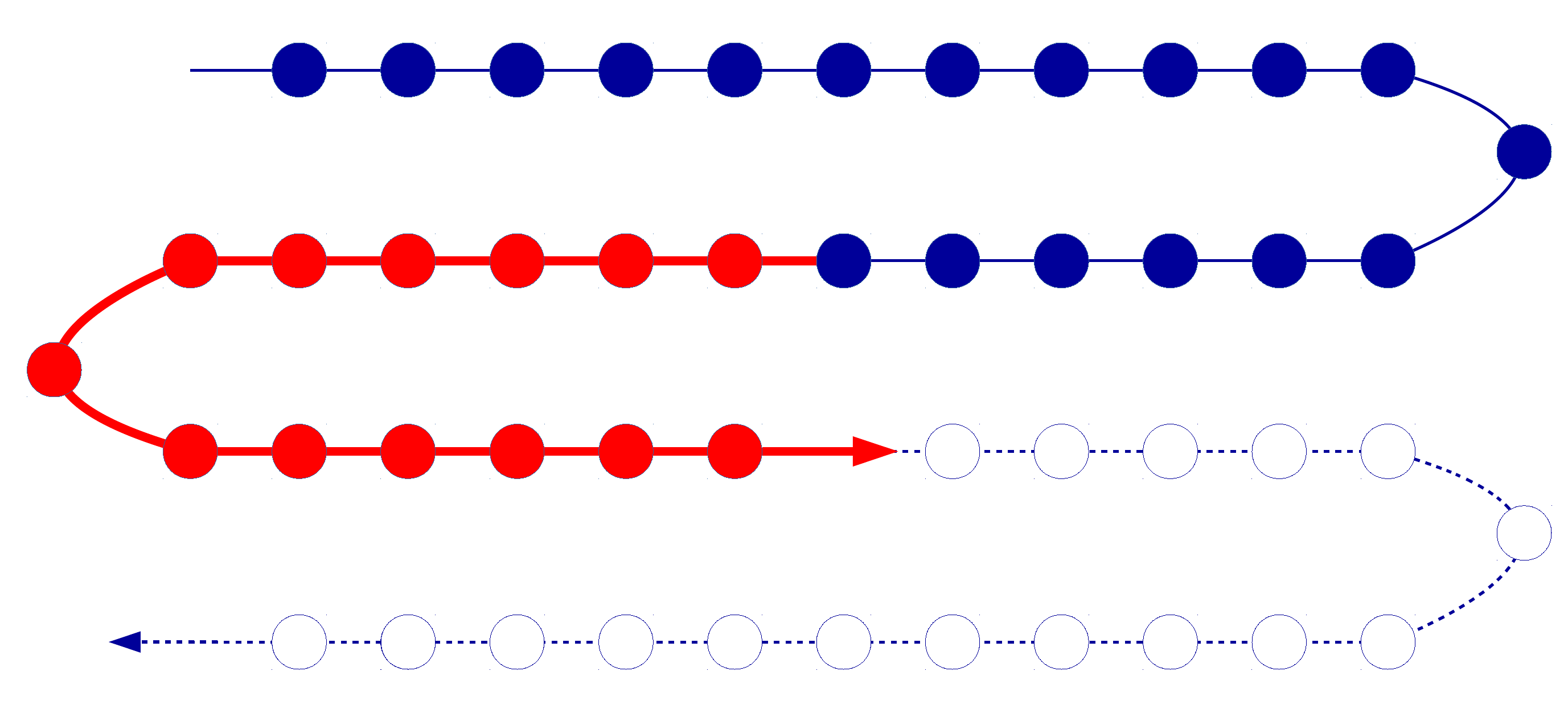}
\end{subfigure}
\caption{(Left) Illustration of an edge connecting two nodes $x_i$ and $x_j$. The error $e_{i,j}$ is computed between the measurement $z_{i,j}$ and the predicted measurement $\hat{z}_{i,j}$. In addition, each edge encodes an information matrix $\Omega_{i,j}$ that represents the uncertainty of the measurement; (Right) Sliding-window sub-graph optimization: nodes that belong to the current sub-graph are painted in red, old nodes no more optimized are painted in blue, while nodes that will be added in the future are painted in white.}\label{fig:edge_scheme_sliding_window}
\end{figure}

\begin{figure}[t!]
		\centering
		\includegraphics[width=\linewidth]{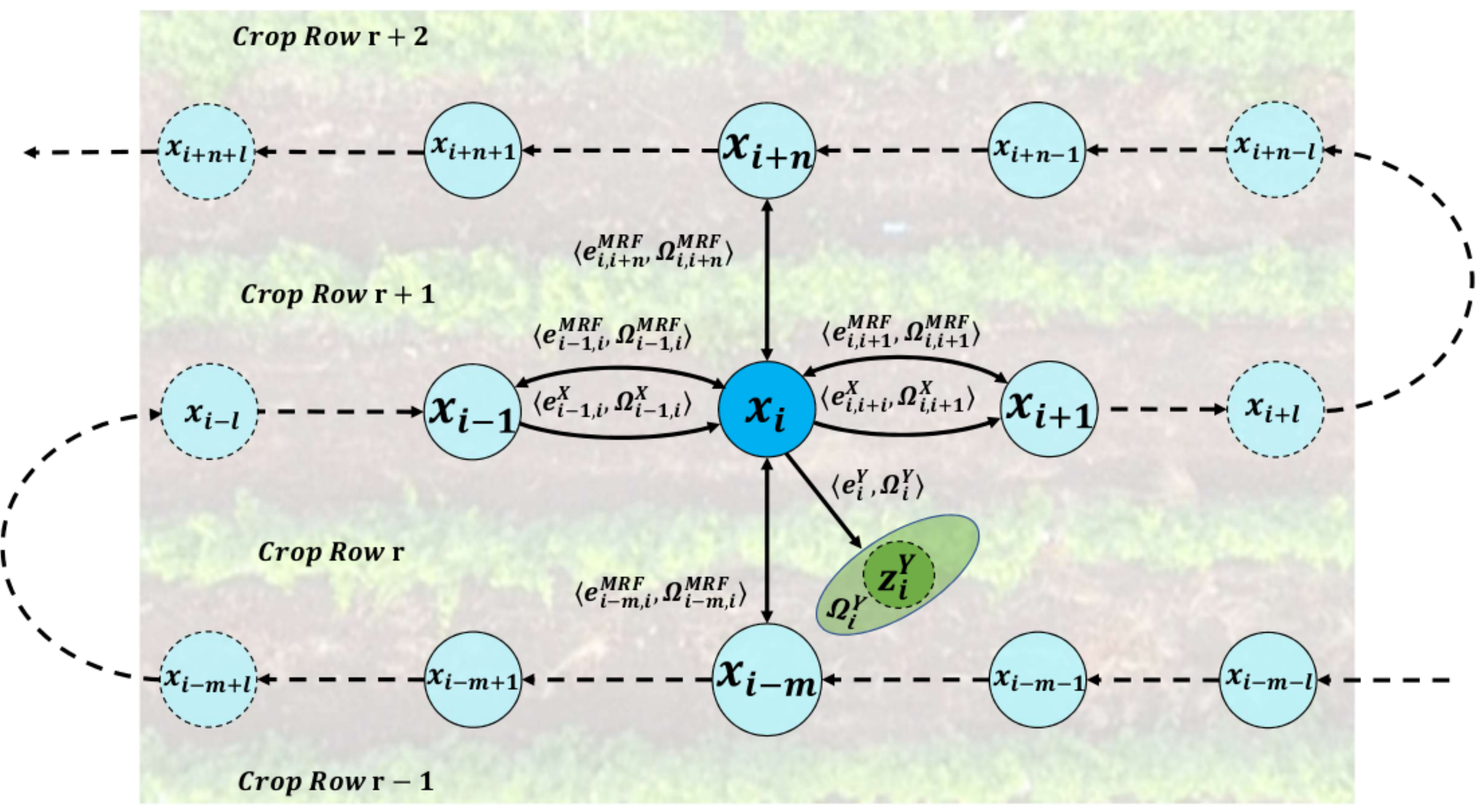}
		\caption{Overview of the built pose graph. Solid arrows represent graph edges, that encode conditional dependencies between nodes, dotted arrows temporal relationships between nodes. For the sake of clarity, we show here only the edges directly connected with the node $x_i$, by representing only one instance for each class of edges: (i) the binary non directed MRF constraint $\langle e^{MRF}_{i,i+1}, \Omega^{MRF}_{i,i+1} \rangle$; (ii) the binary directed edge $\langle e^{\mathcal{X}}_{i,i+1}, \Omega^{\mathcal{X}}_{i,i+1} \rangle$ induced from sensor $\mathcal{X} \in \{VO, WO, AMM, LID \}$; (iii) the unary edge $\langle e_{i}^{\mathcal{Y}}, \Omega^{\mathcal{Y}}_{i} \rangle$ induced by sensor $\mathcal{Y} \in \{GPS, DEM, IMU\}$. We superimposed the graph on a cultivated field to remark the relationship between the graph structure and the crop rows arrangement.}\label{fig:overall_scheme}
\end{figure}

The proposed system handles the global pose estimation problem as a pose graph optimization problem. A pose graph is a special case of factor graph\footnote{A factor graph is a bipartite graph where nodes encode either variables or measurements, namely the factors.}, where the factors $\langle \cdot \rangle$ are only connected to variables (i.e., nodes) pairs, and variables are only represented by robot poses. For this reason, it is common to represent each factor with an edge. Solving a factor graph means finding a configuration of the nodes for which the likelihood of the actual measurements is maximal. Since we assume that all the involved noises follow a Gaussian distribution, we can solve this problem by employing an iterative least square approach.\\
We define $X = \{x_0, . . . , x_{N-1}\}$ as the vector of graph nodes that represents the robot poses at discrete points in time, where each $x_i=(T_i,R_i)$ is represented by the full 3D pose in terms of a translation vector $T_i = [t_{x,i}~t_{y,i}~t_{z,i}]'$ and, using the axis-angle representation, an orientation vector $R_i = [r_{x,i}~r_{y,i}~r_{z,i}]'$, both in $\mathbb{R}^3$. This pose is defined with respect to a global reference centered in $x_0$\footnote{We transform each global measurement (e.g., GPS measurements) in the reference frame $x_0$.}. We denote with $z$ the sensor measurements that can be related to pairs or single nodes.  Let $z_{i,j}$ be a relative motion measurement between nodes $x_i$ and $x_j$, while $z_{i}$ be a global pose measurement associated to the node $x_i$. Additionally, let $\Omega_{i,j}$ and $\Omega_{i}$ represent the information matrices encoding the reliability of such measurements, respectively.
From the poses of two nodes $x_i$ and $x_j$, it is possible to compute the expected relative motion measurement $\hat{z}_{i,j}$ and the expected global measurement $\hat{z}_i$ (see Fig.~\ref{fig:edge_scheme_sliding_window}, left). We formulate the errors between those quantities as:
\begin{equation}
 e_{i,j} = z_{i,j} - \hat{z}_{i,j}, ~ e_{i} = z_{i} - \hat{z}_{i},
 \label{eq:errors}
\end{equation}
Thus, for a general sensor $\mathcal{X}$ providing a relative information, we can characterize an edge (i.e., a \textit{binary factor} $\langle e^{\mathcal{X}}_{i,i+1}, \Omega^{\mathcal{X}}_{i,i+1} \rangle$) by the error $e^{\mathcal{X}}_{i,i+1}$ and the information matrix $\Omega^{\mathcal{X}}_{i,j}$ of the measurement, as described in \cite{grisetti10titsmag}. In other words, an edge represents the relative pose constraint between two nodes (Fig.~\ref{fig:edge_scheme_sliding_window}, left).
In order to take into account also global pose information, we use unary constraints, namely a measurement that constrains a single node. Hence, for a general sensor $\mathcal{Y}$ providing an absolute information, we define $\langle e_{i}^{\mathcal{Y}}, \Omega^{\mathcal{Y}}_{i} \rangle$ as the prior edge (i.e., an \textit{unary factor}) induced by the sensor $\mathcal{Y}$ on node $x_i$. Fig.~\ref{fig:overall_scheme} depicts a portion of a pose graph highlighting both unary and binary edges. Each edge acts as a directed spring with elasticity inversely proportional to the relative information matrix associated with the measurement that generates the link. Our pose graph is built by adding an edge for each sensor reading, for both relative (e.g., wheel odometry readings) and global (e.g., GPS readings) information. In addition, we propose to integrate other prior information that exploit both the specific target environment and the assumptions we made. In the following, we report the full list of edges exploited in this work, divided between local (relative) and global measurements (we report in brackets the acronyms used in Fig.~\ref{fig:overall_scheme}):

\textbf{Local measurements}: Wheel odometry measurements (WO), Visual odometry estimations (VO), Elevation constraints between adjacent nodes (MRF), Ackermann motion model (AMM), Point-clouds local registration (LID). 

\textbf{Global measurements}: GPS readings (GPS), Digital Elevation Model data (DEM), IMU readings (IMU).

We define $\langle e^{VO}_{i,i+1}, \Omega^{VO}_{i,i+1} \rangle$ as the relative constraint induced by a visual odometry algorithm, $\langle e^{WO}_{i,i+1}, \Omega^{WO}_{i,i+1} \rangle $ as the relative constraint induced by the wheel odometry, and $\langle e^{LID}_{i,i+1}, \Omega^{LID}_{i,i+1} \rangle$ as the relative constraint obtained by aligning the local point-clouds perceived by the 3D LIDAR sensor. 

Often, GPS and visual odometry provide poor estimates of the robot position along the $z$-axis (i.e, the axis that represents its \textit{elevation}). In the GPS case, this low accuracy is mainly due to the Dilution of Precision, multipath or atmospheric disturbances, while in the visual odometry this is due to the 3D locations of the tracked points. In a typical agricultural scenario most of the visual features belong to the ground plane. Hence, the small displacement of the features along the z-axis may cause a considerable drift. On the other hand, agricultural fields usually present locally flat ground levels and, moreover, a farming robot usually traverses the field by following the crop rows. Driven by these observations, we enforce the local ground smoothness assumption by introducing an additional type of local constraints that penalizes the distance along the $z$-coordinate between adjacent nodes. Therefore, the built pose graph can be augmented by a 4-connected MRF \cite{blake2011markov}: each node is conditionally connected with the previous and the next nodes in the current crop row, and with the spatially closest nodes that belong to the previous and next crop rows, respectively. We refer to this constraint as $\langle e^{MRF}_{i,i+1}, \Omega^{MRF}_{i,i+1} \rangle$ in Fig.~\ref{fig:overall_scheme} (e.g., the set $\{ x_{i-1}, x_i, x_{i+1}, x_{i-m}, x_{i+n} \}$). We then add a further type of local constraint based on the Ackermann steering model, that assumes that the robot is moving on a plane. In this work, we relax this assumption to local planar motions between temporal adjacent nodes. Such a motion plane is updated with the attitude estimation of the subsequent node. We integrate this constraint by means of a new type of edge, namely $\langle e^{AMM}_{i,i+1}, \Omega^{AMM}_{i,i+1} \rangle$.

Local constraints are intrinsically affected by a small cumulative drift: to overcome this problem, we integrate in the graph drift-free global measurements as position prior information. In particular, we define a GPS prior $z^{GPS}_i$ and an IMU prior $z^{IMU}_i$ with associated information matrices $\Omega^{GPS}_i$ and $\Omega^{IMU}_i$. The IMU is used as a drift-free roll and pitch reference\footnote{We experienced that integrating the full inertial information inside the optimization did not positively affect the state estimation: our intuition is that the slow, often unimodal, motion of our robot makes the IMU biases difficult to estimate and sometimes predominant over the motion components.}, where the drift resulting from the gyroscopes integration is compensated by using the accelerometers data.

Finally, we introduce an additional global measurement by means of an altitude prior, provided by a DEM. A DEM is a special type of Digital Terrain Model that represents the elevation of the terrain at some location, by means of a regularly spaced grid of elevation points \cite{Hirt2014}. The DEM maps a 2D coordinate to an absolute elevation. Since we assume that the altitude varies slowly, we can use the current position estimate $T_i$ (i.e., the $t_{x,i}$ and $t_{y,i}$ components) to query the DEM for a reliable altitude estimation $z_{DEM,i}= f(t_{x,i}, t_{y,i})$, with associated information matrix $\Omega^{DEM}_i$. The cost function is then assembled as follows:

\begin{eqnarray}
 J_i =  \sum_{i=1}^{N-1} \Big( \underbrace{\sum_{\mathcal{X}}  e_{i,i-1}^\mathcal{X}\Omega_{i,i-1}^\mathcal{X}e_{i,i-1}^\mathcal{X'}}_{Binary\ constraints} + 
       \underbrace{\sum_{\mathcal{Y}}  e_i^\mathcal{Y}\Omega_i^{\mathcal{Y}} e_i^\mathcal{Y'}}_{Unary\ constraints} + \nonumber\\ \underbrace{ \sum_{j\in \mathbb{N}_i} e_{i,j}^{MRF}\Omega_{i,j}^{MRF}e_{i,j}^{MRF'}}_{MRF\ constraint}\Big) 
 \label{eq:cost_function}
\end{eqnarray}

where $\mathcal{X}$ and $\mathcal{Y}$ represent respectively the set of binary and unary constraints defined above (see Fig.~\ref{fig:overall_scheme}), and $\mathbb{N}_i$ stands for the 4-connected neighborhood of the node $x_i$. 

\section{Pose Graph Optimization}\label{sec:graph_optimization}

In this section, we focus on the solution of the cost function reported in Eq.~\ref{eq:cost_function}, describing the error computation, the weighting factors assignment procedure and the on-line and off-line versions of the optimization. We finally report some key implementation insights.

\subsection{Error Computation}

For each measurement $z$, given the current graph configuration, we need to provide a prediction $\hat{z}$ in order to compute errors in Eq.~\ref{eq:cost_function}. $\hat{z}$ represents the expected measurement, given a configuration of the nodes, which are involved in the constraint. Usually, for a binary constraint, this prediction is the relative transformation between the nodes $x_i$ and $x_j$, while for an unary constraint it is just the full state $x_i$ or a subset of its components.
We define $\textbf{X}_i$ as a general homogeneous transformation matrix related to the full state of the node $x_i$ (e.g., the homogeneous rigid body transformation generated from $T_i$ and $ R_i$) and $\varPhi(\cdot)$ as a generic mapping function from $\textbf{X}_i$ to a vector; now, we can express $\hat{z}_{i,j}$ and $\hat{z}_i$ as:
\begin{equation}
\hat{z}_{i,j} =  \varPhi( \textbf{X}^{-1}_i \cdot \textbf{X}_j ) , ~ \hat{z}_i = \varPhi( \textbf{X}_i )
\label{eq:error_composition}
\end{equation}
In this work not all the constraints belong to $SE(3)$: indeed, most of used sensors (e.g., WO, IMU) can only observe a portion of the full state encoded in $x$. Therefore, in the following, we will show how we obtain the expected $\hat{z}$ for each involved cue (for some components, we omit the subscripts $i$ and $j$ by using the relative translations $dt$ and rotations $dr$ between adjacent nodes):

\textbf{VO and LID}: these front-ends provide the full 6D motion: we build $\hat{z}^{VO}$ and $\hat{z}^{LID}$ by computing the relative transformation between the two connected nodes as in Eq.~\ref{eq:error_composition};

\textbf{WO}: the robot odometry provides the planar motion by means of a roto-translation $z_{WO} = (dt_x, dt_y, dr_z)$: we build $\hat{z}^{WO}$ as  $\varPhi( \textbf{X}^{-1}_i \cdot \textbf{X}_j )|_{t_x,t_y,r_z}$, the subscripts after $\varPhi(\cdot)$ specify that the map to the vector $\hat{z}$ involves only such components; 

\textbf{MRF and DEM}: they constrain the altitude of the robot, we obtain the estimated measurements as: 

\begin{subequations}
\begin{align}
        \hat{z}^{MRF}_{i,j} = (0,0,t_{z,i} - t_{z,j},0,&0,0)\label{eq:mrf_z}\\
        \hat{z}^{DEM}_i = (0,0,t_{z,i},0,0,0)\label{eq:dem_z}
\end{align}
\label{eq:dem_mrf_z}
\end{subequations}

\textbf{GPS}: this sensor only provides the robot position:
\begin{equation}
 \hat{z}^{GPS}_i = (T_i, 0_{3\times1})
\end{equation}

\textbf{IMU}: from this measurement we actually exploit only the \textit{roll} and \textit{pitch} angles, being the rotation around the $z$ axis provided by the IMU usually affected by not negligible inaccuracies. Therefore, we obtain $\hat{z}^{IMU}_i = \varPhi( \textbf{X}_i )|_{r_{x,i},r_{y,i}}$;

\textbf{AMM}: we formulate such a constraint by a composition of two transformation matrices. The first one encodes a roto-translation of the robot around the so called \textit{Instantaneous Center of Rotation} (ICR). We follow the same formulation presented in \cite{pretto2011}:

\begin{equation}
    \textbf{X}(\rho, dr_z) =
  \begin{bmatrix}
    cos(\frac{dr_z}{2}) & -sin(\frac{dr_z}{2}) & 0 & \rho \cdot cos(\frac{dr_z}{2}) \\
    sin(\frac{dr_z}{2}) & cos(\frac{dr_z}{2})  & 0 & \rho \cdot sin(\frac{dr_z}{2}) \\
    0 & 0 & 1 & 0 \\
    0 & 0 & 0 & 1 \\
  \end{bmatrix}
 \label{eq:I_part_ackermann}
\end{equation}

where $\rho$ is the norm of the translation along $dt_x$ and $dt_y$. Additionally, we add a further rotation along those two axes, taking also into account the ground slope, by rotating the ideal plane on which the vehicle steers following the Ackermann motion model:

\begin{equation}
    \textbf{X}(dr_x, dr_y) =
  \begin{bmatrix}
    R(dr_x,dr_y) & 0_{1x3} \\
    0_{3x1} & 1
  \end{bmatrix}
 \label{eq:II_part_ackermann}
\end{equation}

Hence, we obtain $\hat{z}^{AMM}$ as $\varPhi( \textbf{X}(dr_x, dr_y) \cdot \textbf{X}(\rho, dr_z) )$.

\subsection{Dynamic Weight Assignment}

The impact of each constraint in the final cost function (Eq.~\ref{eq:cost_function}) is weighted by its relative information matrix. As a consequence, such information matrices play a crucial role in weighting the measurements, i.e. giving much reliability to a noisy sensor can lead to errors in the optimization phase. We tackle this problem by dynamically assigning the information matrix for each component as follows:

\textbf{WO}: we use as information matrix $\Omega^{WO}_{i,j}$ the inverse of the covariance matrix $\Sigma^{WO}$ of the robot odometry, scaled by the magnitude of the distance and rotation traveled between the nodes $x_i$ and $x_j$, as explained in \cite{Thrun2005};

\textbf{VO}: we use the inverse of the covariance matrix $\Sigma^{VO}$ provided as output by the visual odometry front-end, weighting the rotational and translational sub-matrices ($\Sigma^{VO,R}$ and $\Sigma^{VO,T}$) with two scalars $\lambda_{VO,R}$ and $\lambda_{VO,T}$, experimentally tuned. Since we do not directly tune the VO system internal parameters, we employ these ''VO agnostic`` scaling factors that have the analogous effects as injecting a higher sensor noise. In the experiments, we set $\lambda_{VO,R}=5$ and $\lambda_{VO,T}=1$;

\textbf{MRF}: we set the information matrix $\Omega^{MRF}_{i,j} = \textit{diag}(0,0,w^{MRF}_z,0,0,0)$. The weight $w^{MRF}_z = \lambda_{MRF}/|x_i - x_j|_{t_x,t_y}$ is inversely proportional to the distance in the $(x,y)$ plane between the two nodes, while $\lambda_{MRF}$ has been experimentally tuned. $\lambda_{MRF}=0.8$ in the experiments;

\textbf{GPS}: we use as information matrix $\Omega^{GPS}_{i}$, the inverse of the covariance matrix $\Sigma^{GPS}$ provided by the GPS sensor;

\textbf{AMM}: we use as information matrix $\Omega^{AMM}_{i,j}$, an identity matrix scaled by the magnitude of the traveled distance between the nodes $x_i$ and $x_j$, similarly to the wheel odometry constraint. This allows to model the reliability of such a constraint as inversely proportional to the traveled distance;

\textbf{IMU}: we use as information matrix $\Omega^{IMU}_{i}$, the inverse of the covariance matrix $\Sigma^{IMU}$ provided by the IMU sensor;

\textbf{DEM}: we set the information matrix $\Omega^{DEM}_{i} = diag(0,0,w^{DEM}_z,0,0,0)$, where $w^{DEM}_z$ is empirically tuned. In the experiments we set $w^{DEM}_z=5$;

\textbf{LID}: we set the information matrix $\Omega^{LID}_{i,j}$ as the inverse of the covariance matrix estimated from the transformation provided by the registration algorithm (e.g., an ICP algorithm), by using the procedure described in \cite{Prakhya_2015}. Such an information matrix allows adapting the influence of the point-cloud alignment inside the optimization process, enabling to correctly deal also with the lack of geometrical structure on some dimensions, e.g. in farming scenarios with small plants.

\subsection{Sliding-Window Optimization}\label{sec:sliding_window}

A re-optimization of the whole pose graph presented above, every time a new node is added, cannot guarantee the real-time performances required for on-line field operations, especially when the graph contains a large amount of nodes and constraints. We solve this issue by employing a sliding-window approach, namely performing the optimization procedure only on a sub-graph that includes a sequence of recent nodes. 
Each time a new node associated with the most recent sensor readings is added to the graph, the sub-graph is updated by adding the new node and removing its oldest one, in a SW fashion. The optimization process is performed only on the current sub-graph, while older nodes maintain the state assigned during the last optimization where they were involved. In order to preserve the MRF constraints, the size of the sub-graph is automatically computed so that any adjacent nodes in the previous row are included (see Fig. \ref{fig:edge_scheme_sliding_window}, right). A global optimization of the whole pose graph is then carried out off-line, using as initial guess the node states assigned on-line using the SW approach.

\subsection{Implementation Details}

\textbf{Temporal Synchronization}: In the previous sections, we tacitly assumed that all sensor measurements associated with a graph node share the same time stamp. However, in a real context, this is usually not true. In our implementation, we trigger the creation of new nodes every $step_{WO}$ meters ($0.3~m$ in our experiments), by using the wheel odometry as a distance reference. We associate to each node synchronized estimates of the other sensor readings, obtained by means of linear interpolation over the closest readings of each used sensor. This enables to associate to the same node a set of heterogeneous sensor readings that share the same time stamp.\\
\textbf{Visual Odometry Failures}: 
VO systems are usually tuned by default to provide high accuracy at the expense of the robustness. We address this limitation by employing a simple strategy designed to mitigate VO failures. We exploit the local reliability of the WO: when the difference between WO and VO is greater than a given threshold, we assume a failure of the latter. In this case, we reduce the influence of the VO during the pose graph optimization by downscaling its information matrix.\\
\textbf{Point-Cloud Registration}: Point-clouds acquired by a 3D LIDAR are typically too sparse to perform a robust alignment: thus, we accumulate a number of LIDAR readings into a single point-cloud by using the motion estimations provided by the VO. The point-cloud registration is finally performed using the Iterative Closest Point (ICP) algorithm.\\
\textbf{Graph Optimization}: We perform both the on-line and off-line pose graph optimizations (Sec.~\ref{sec:sliding_window}) using the Levenberg-Marquardt algorithm implemented in the \textit{g2o} graph optimization framework \cite{kummerle2011}.

\section{Experiments}\label{sec:experiments}

In order to analyze the performance of our system, we collected two datasets\footnote{\texttt{www.dis.uniroma1.it/{\texttildelow}labrococo/fsd}} with different UGV steering modalities. In \textit{Dataset A} the robot follows 6 adjacent crop rows by constantly maintaining the same global orientation, e.g. half rows have been traversed by moving the robot backward, while in \textit{Dataset B} the robot is constantly moving in the forward direction.
Both datasets include data from the same set of sensors: (i) wheel odometry; (ii) VI-Sensor device (stereo camera + IMU) \cite{nikolic2014}; (iii) Velodyne VLP-16 3D LIDAR; (iv) a low cost U-blox RTK-GPS; (v) an U-blox Precise Point Positioning (PPP) GPS. For a comprehensive description of the UGV farm robot, the sensors setup and the calibration procedure, we refer the interested readers to the on-line supplementary material\footnote{\texttt{www.dis.uniroma1.it/{\texttildelow}labrococo/fsd/ral2018sup.pdf}\label{suppl_mat}}.\\
In all our experiments, we employ Stereo DSO~\cite{wang2017stereo} as VO subsystem and the ICP implementation provided by the PCL library as point-cloud registration front-end. The IMU, the wheel odometry and both the GPSs provide internally filtered outputs (attitude, relative and absolute positions, respectively), along with covariance matrices associated to the outputs in the IMU and GPSs cases. We built the DEM of the inspected field by using the Google Elevation API that provides, for the target field, measurements over a regularly spaced grid with a resolution of 10 meters. We interpolated such measurements to provide a denser information.
We acquired a ground truth 3D reference by using a LEICA laser tracker. This sensor tracks a specific target mounted on the top of the robot and provides a position estimation ($x$, $y$ and $z$) with millimeter-level accuracy.
Both datasets have been acquired by using the Bosch BoniRob farm robot (Fig. \ref{fig:teaser}, left) on a field in Eschikon, Switzerland (Fig. \ref{fig:teaser}, right).
In addition to these two datasets, we have created a third dataset (\textit{Dataset C}), where we simulate a sudden RTK-GPS signal loss, e.g. due to a communication loss between the base and the rover stations. In particular, we simulated the accuracy losses by randomly switching for some time to the PPP-GPS readings. 

In the following, we report the quantitative results by using the following statistics build upon the localization errors with respect to the ground truth reference: Root Mean Square Error ($RMSE$ in the tables), maximum and mean absolute error ($Max$ and $Mean$), and mean absolute error along each component ($err_x$, $err_y$ and $err_z$).

\subsection{Dataset A and Dataset B}
\label{sec:datA_and_datB}
\begin{table*}[ht!]
	\scriptsize
	\centering
	\caption{Error statistics in \textit{Dataset A} and \textit{Dataset B} by using different sensor setups and constraints for the global, off-line and the sliding-window (SW), on-line pose graph optimization procedures. The results of the ORB SLAM 2 system (OS2 in the table) are reported for both type of GPSs.}
	\label{tab:err_statistic_datasetA}

	  \begin{tabular}{ ccccccccccccccccccc }
	  \multicolumn{9}{c}{} & \multicolumn{5}{c}{ \textbf{DatasetA} } &\multicolumn{5}{c}{ \textbf{DatasetB} }\\
	  \cmidrule(lr){10-14} \cmidrule(lr){15-19}
	  \rotatebox[origin=c]{90}{\parbox[c]{1cm}{\centering GPS}}&\rotatebox[origin=c]{90}{\parbox[c]{1cm}{\centering WO}} &\rotatebox[origin=c]{90}{\parbox[c]{1cm}{\centering VO}} &\rotatebox[origin=c]{90}{\parbox[c]{1cm}{\centering IMU}} 
	  &\rotatebox[origin=c]{90}{\parbox[c]{1cm}{\centering AMM}} &\rotatebox[origin=c]{90}{\parbox[c]{1cm}{\centering ELEV}}&\rotatebox[origin=c]{90}{\parbox[c]{1cm}{\centering LIDAR}}
	  &\rotatebox[origin=c]{90}{\parbox[c]{1cm}{\centering MRF}} &\rotatebox[origin=c]{90}{\parbox[c]{1cm}{\centering SW}} &$err_x$ &$err_y$ &$err_z$ &$Max$ &$RMSE$ &$err_x$ &$err_y$ &$err_z$ &$Max$ &$RMSE$ \\\cline{1-19}\\
	  \multirow{15}{*}{\rotatebox[origin=c]{90}{\parbox[c]{1cm}{\centering PPP}}}
	  
	  &&&&&&&& 											 &0.349  &0.582  &1.577  &2.959    &1.710  &0.306  &0.501  &1.484  &2.875    &1.621 \\
	  &\checkmark&&&&&&& 										 &0.311  &0.520  &1.537  &2.954    &1.630  &0.246  &0.416  &1.424  &2.829    &1.504 \\
	  &\checkmark&&&&\checkmark&&&                                 					 &{0.343} &{0.572} &{0.475} &{1.627} &{1.071} &{0.241} &{0.408} &{0.492} &{1.782} &{1.168} \\
	  &\checkmark&\checkmark&&&&&& 									 &{0.239}  & {0.412}  & {0.672}  & {1.628}  & {0.961}    & {0.222}  & {0.362}  & {1.298}  & {2.392}   & {1.211} \\
	  &\checkmark&\checkmark&&&&\checkmark&& 							 &{0.233} &{0.422} &{0.649} &{1.421} &{0.863} &{0.227} &{0.361} &{1.292} &{2.571} &{1.242} \\
	  &\checkmark&\checkmark&&&&&\checkmark& 							 &{0.239} &{0.411} &{0.528} &{1.398} &{0.719} &{0.221} &{0.364} &{1.019} &{2.362} &{1.119} \\
	  &\checkmark&\checkmark&&\checkmark&&\checkmark&& 						 &{0.224} &{0.411} &{0.551} &{1.375} &{0.726} &{0.201} &{0.397} &{0.881} &{2.019} &{0.951} \\
	  &\checkmark&\checkmark&\checkmark&&&\checkmark&& 						 &{0.222} &{0.389} &{0.531} &{1.281} &{0.729} &{0.229} &{0.407} &{0.652} &{1.613} &{0.829} \\
	  &\checkmark&\checkmark&\checkmark&\checkmark&&&&			 			 &{0.239} &{0.361} &{0.523} &{1.272} &{0.739} &{0.231} &{0.369} &{0.641} &{1.461} &{0.732} \\
	  &\checkmark&\checkmark&\checkmark&\checkmark&&\checkmark&& 					 &{0.224} &{0.371} &{0.453} &{1.124} &{0.621} &{0.221} &{0.362} &{0.619} &{1.611} &{0.734} \\
	  &\checkmark&\checkmark&\checkmark&\checkmark&&&\checkmark& 					 &{0.234} &{0.360} &{0.440} &{1.093} &{0.564} &{0.199} &{0.360} &{0.475} &{1.161} &{0.660} \\
	  &\checkmark&\checkmark&\checkmark&\checkmark&\checkmark&\checkmark&& 				 &{0.234} &{0.342} &{0.311} &{0.921} &{0.422} &{0.198} &{0.361} &{0.463} &{1.121} &{0.604} \\
	  &\checkmark&\checkmark&\checkmark&\checkmark&\checkmark&&\checkmark& 				 &{0.211} &\textbf{0.331} &{\textbf{0.282}} &{0.897} &{0.416} &{0.182} &{0.339} &{0.369} &{1.198} &{0.471} \\
	  &$\pmb{\checkmark}$&$\pmb{\checkmark}$&$\pmb{\checkmark}$&$\pmb{\checkmark}$&$\pmb{\checkmark}$&$\pmb{\checkmark}$&$\pmb{\checkmark}$& &{\textbf{0.201}} &\textbf{0.331} &{0.289} &{\textbf{0.824}} &{\textbf{0.401}} &{\textbf{0.173}} &{\textbf{0.331}} &{\textbf{0.321}} &{\textbf{1.117}} &{\textbf{0.461}} \\ 
	  &\checkmark&\checkmark&\checkmark&\checkmark&\checkmark&\checkmark&\checkmark&\checkmark 	 &{0.252} &{0.419} &{0.349} &{0.991} &{0.549} &{0.291} &{0.431} &{0.459} &{1.291} &{0.652} \\\cline{2-19}\noalign{\vskip 1mm}
	  &OS2+GPS &&&&&&& 	 &{0.234} &{0.417} &{0.643} &{1.534} &{0.915} &{0.209} &{0.401} &{0.371} &{2.123} &{1.047}\\
	  \cline{1-19} \noalign{\vskip 1mm}
	  \multirow{10}{*}{\rotatebox[origin=c]{90}{\parbox[c]{1cm}{\centering RTK}}}
	  &&&&&&&& 										&0.059   &0.051   	   &0.121   &0.431   &0.128   &0.054   &0.062   &0.091   &0.322   &0.122 \\
	  &\checkmark&&&&&&& 									&0.053   &\textbf{0.042}   &0.105   &0.431   &0.125   &0.049   &0.058   &0.086   &0.321   &0.119 \\ 
	  &\checkmark&\checkmark&&&&&& 								&{0.053} &\textbf{0.042}   &{0.054} &{0.279} &{0.088} &{0.047} &{0.048} &{0.062} &{0.192} &{0.091}  \\
	  &\checkmark&\checkmark&\checkmark&&&&& 						&{0.048} &{0.049} 	   &{0.060} &{0.306} &{0.092} &{0.045} &{0.046} &{0.064} &{0.209} &{0.091} \\
	  &\checkmark&\checkmark&&\checkmark&&&& 						&{0.046} &{0.047} 	   &{0.061} &{0.279} &{0.090} &{0.045} &\textbf{0.045} &{0.064} &{0.211} &{0.090} \\
	  &\checkmark&\checkmark&&\checkmark&&\checkmark&& 					&{0.046} &{0.047} 	   &{0.061} &{0.278} &{0.089} &{0.045} &\textbf{0.045} &{0.062} &{0.197} &{0.090} \\
	  &\checkmark&\checkmark&\checkmark&\checkmark&&&& 					&{0.046} &{0.050} 	   &{0.056} &{\textbf{0.248}} &{0.088} &{0.045} &{0.046} &{0.039} &{0.165} &{0.075} \\
	  &\checkmark&\checkmark&\checkmark&&&\checkmark&\checkmark& 				&{0.047} &{0.049} 	   &\textbf{0.034} &{0.251} &{0.076} &{0.045} &{0.046} &{0.035} &{0.154} &{0.074} \\
	  &\checkmark&\checkmark&\checkmark&\checkmark&\checkmark&\checkmark&\checkmark& 	&{0.051} &{0.049} 	   &{0.068} &{0.312} &{0.097} &{0.046} &{0.048} &{0.064} &{0.219} &{0.095} \\
	  &$\pmb{\checkmark}$&$\pmb{\checkmark}$&$\pmb{\checkmark}$&$\pmb{\checkmark}$&&$\pmb{\checkmark}$&$\pmb{\checkmark}$& 	&{\textbf{0.045}} &{0.048} &\textbf{0.034} &{0.260} &{\textbf{0.075}} &{\textbf{0.044}} &{0.046} &{\textbf{0.034}} &{\textbf{0.151}} &{\textbf{0.073}} \\
	  &\checkmark&\checkmark&\checkmark&\checkmark&&\checkmark&\checkmark&\checkmark 	&{0.053} &{0.051} &{0.042} &{0.272} &{0.084} &{0.051} &{0.051} &{0.035} &{0.172} &{0.084}\\\cline{2-19}\noalign{\vskip 1mm}
	  &OS2+GPS &&&&&&& 	&{0.051} &{0.045} &{0.059} &{0.293} &{0.097} &{0.051} &{0.054} &{0.068} &{0.231} &{0.102}\\\hline
	  \end{tabular}	
	  
\end{table*}
\begin{figure}[ht]
  \centering
  \begin{subfigure}{0.49\columnwidth}￼
    \includegraphics[width=\textwidth]{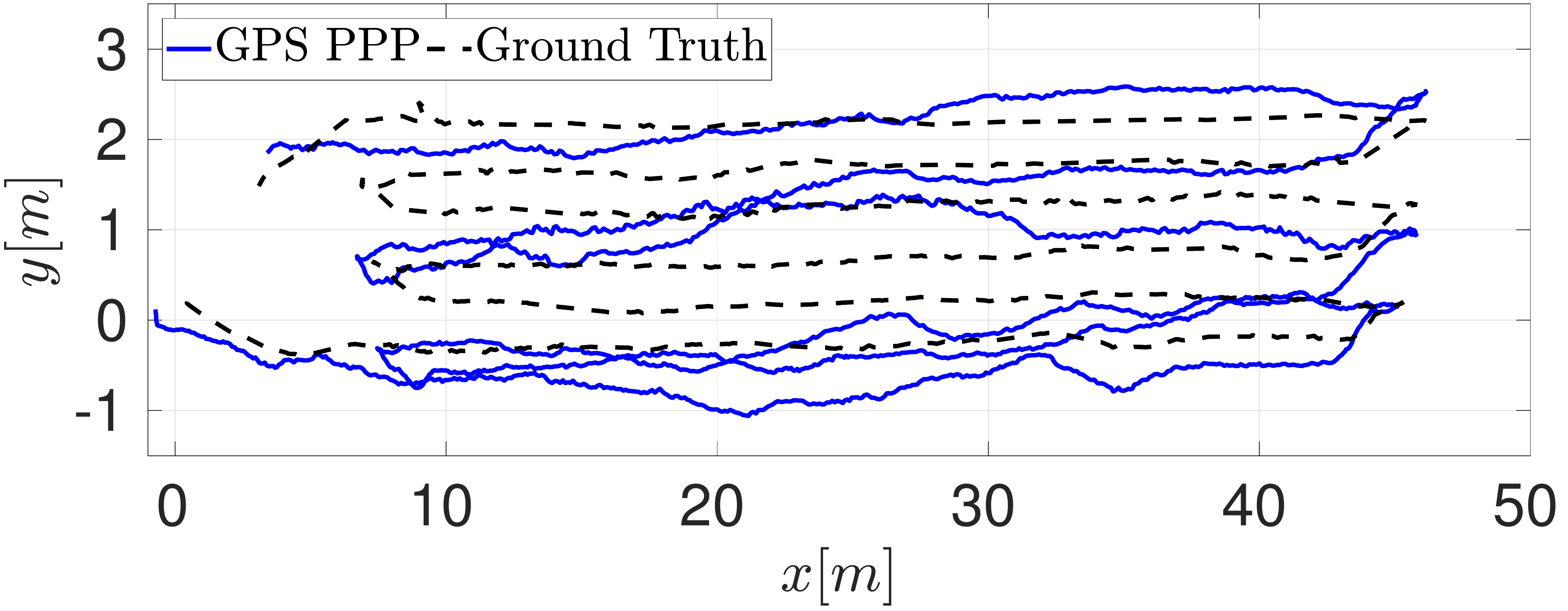}
    \label{fig:gps_ppp_datasetA}
  \end{subfigure}
  \begin{subfigure}{0.49\columnwidth}
    \includegraphics[width=\textwidth]{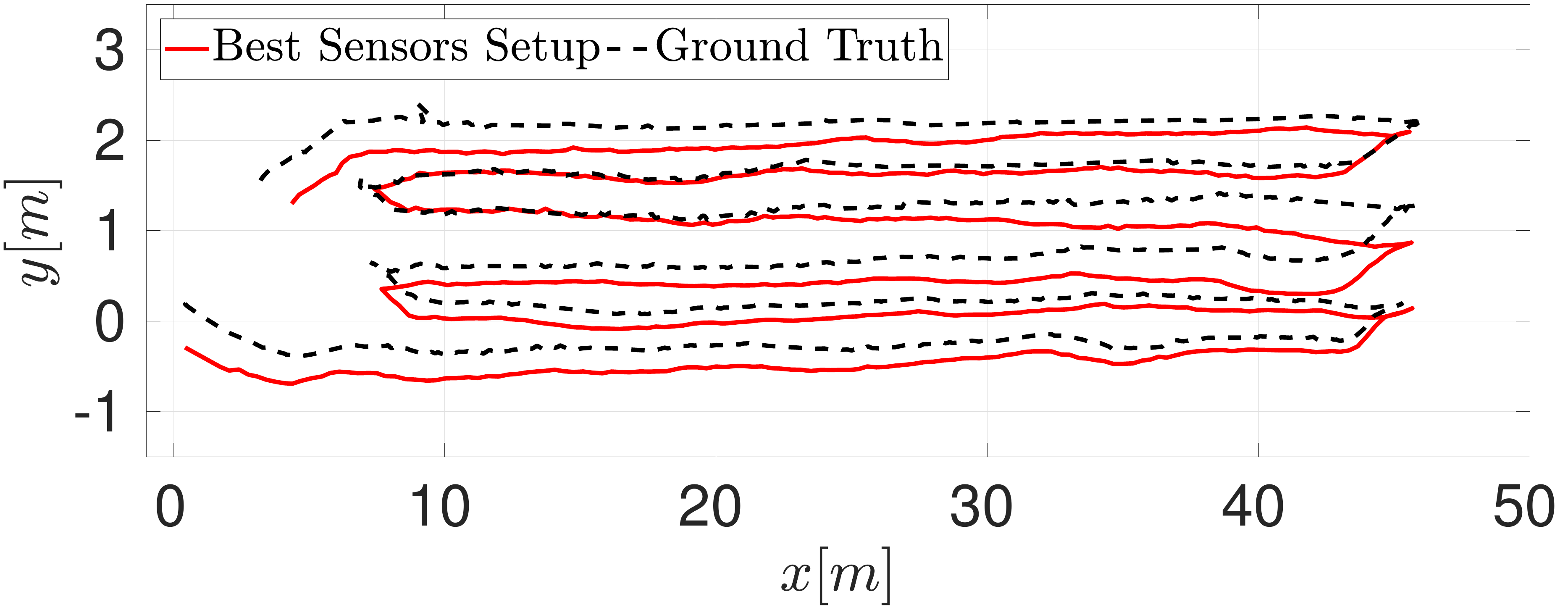}
    \label{fig:gps_best_config_datasetA}
  \end{subfigure}
  \begin{subfigure}{0.49\columnwidth}
    \centering
    \includegraphics[width=\textwidth]{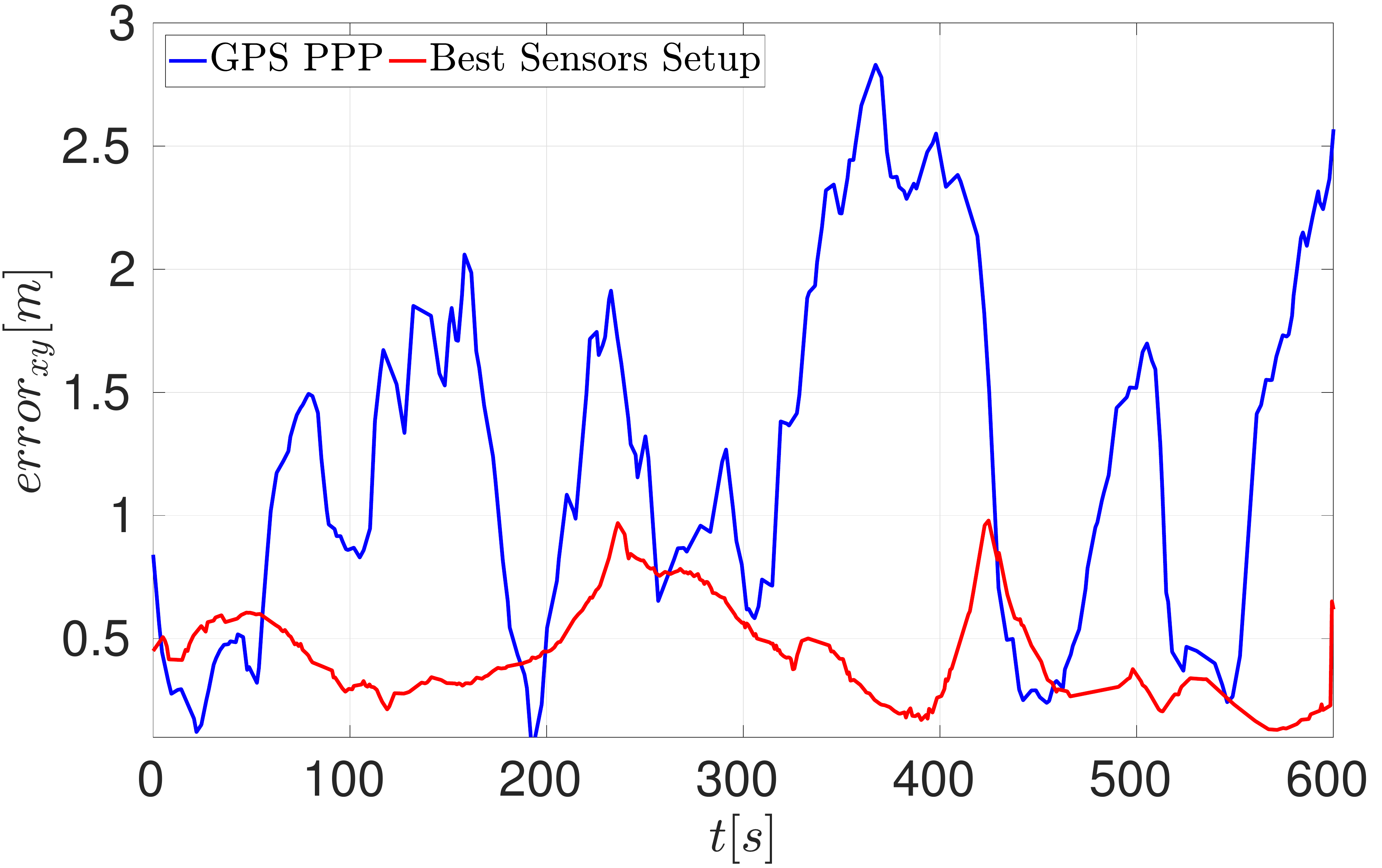}
    \label{fig:errorxy_datasetA}
  \end{subfigure}
  \begin{subfigure}{0.49\columnwidth}
    \centering
    \includegraphics[width=\textwidth]{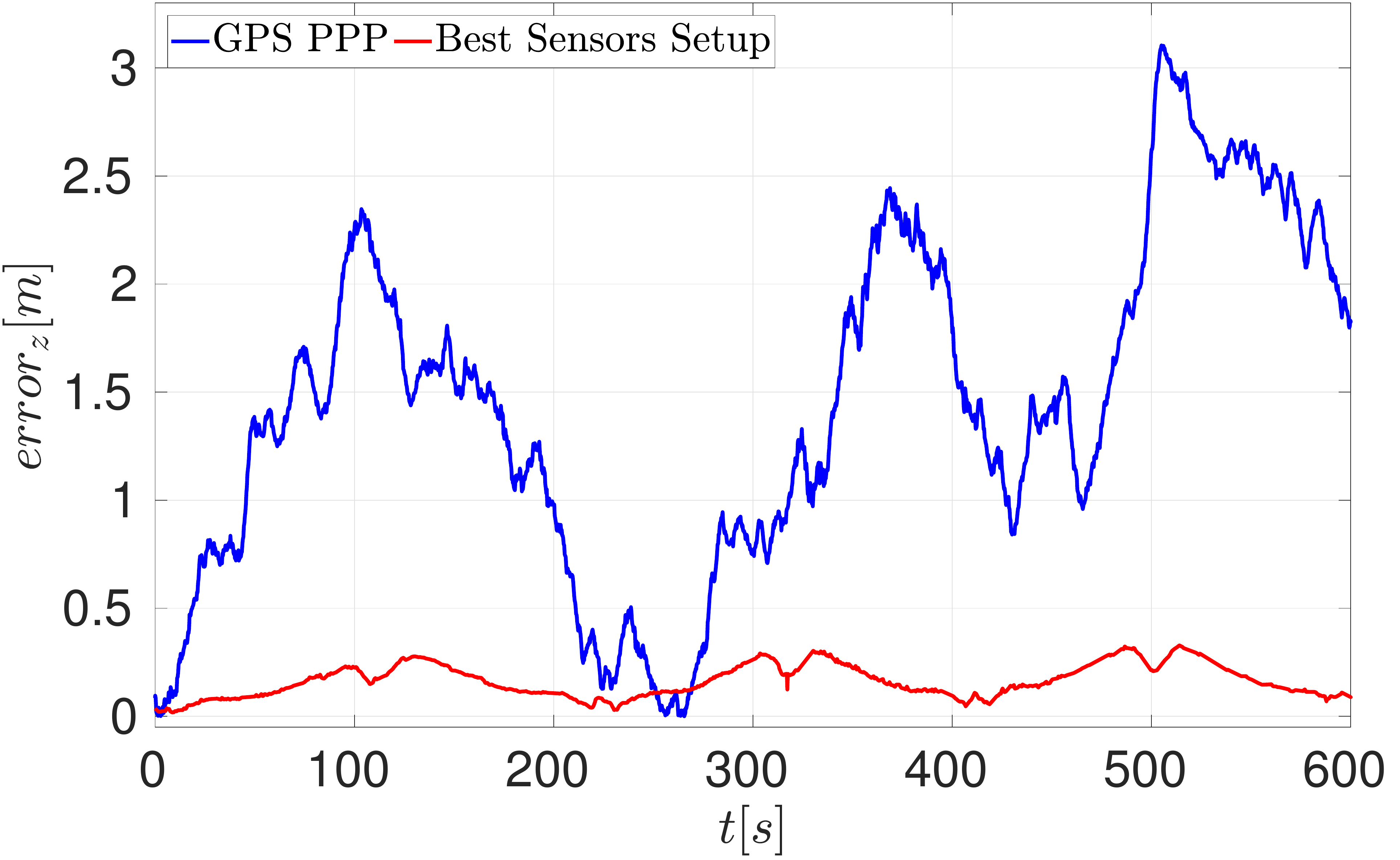}
    \label{fig:errorz_datasetA}
  \end{subfigure}
  \caption{Dataset A, PPP-GPS: (Top) Qualitative top view comparison between the raw GPS trajectory (left) and the optimized trajectory (right); (Bottom): absolute $x,y$ (left) and $z$ (right) error plots for the same trajectories.}
  \label{fig:data_topview_comparison}
\end{figure}
\begin{figure}[h!]
  \centering
  \begin{subfigure}{0.49\columnwidth}
    \includegraphics[width=\textwidth]{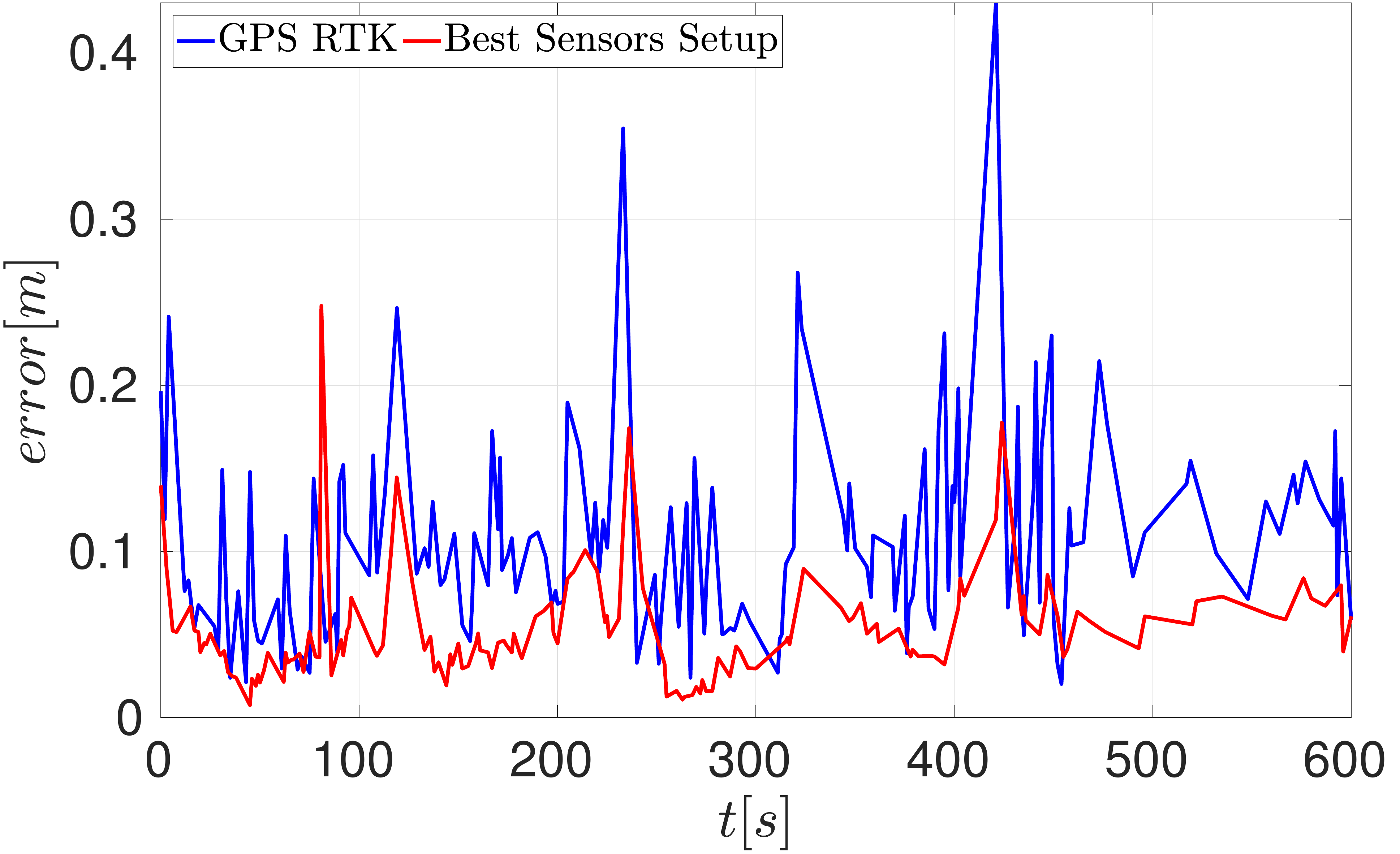}
    \label{fig:datasetC_ppp_data} 
   \end{subfigure}
   \begin{subfigure}{0.475\columnwidth}
    \includegraphics[width=\textwidth]{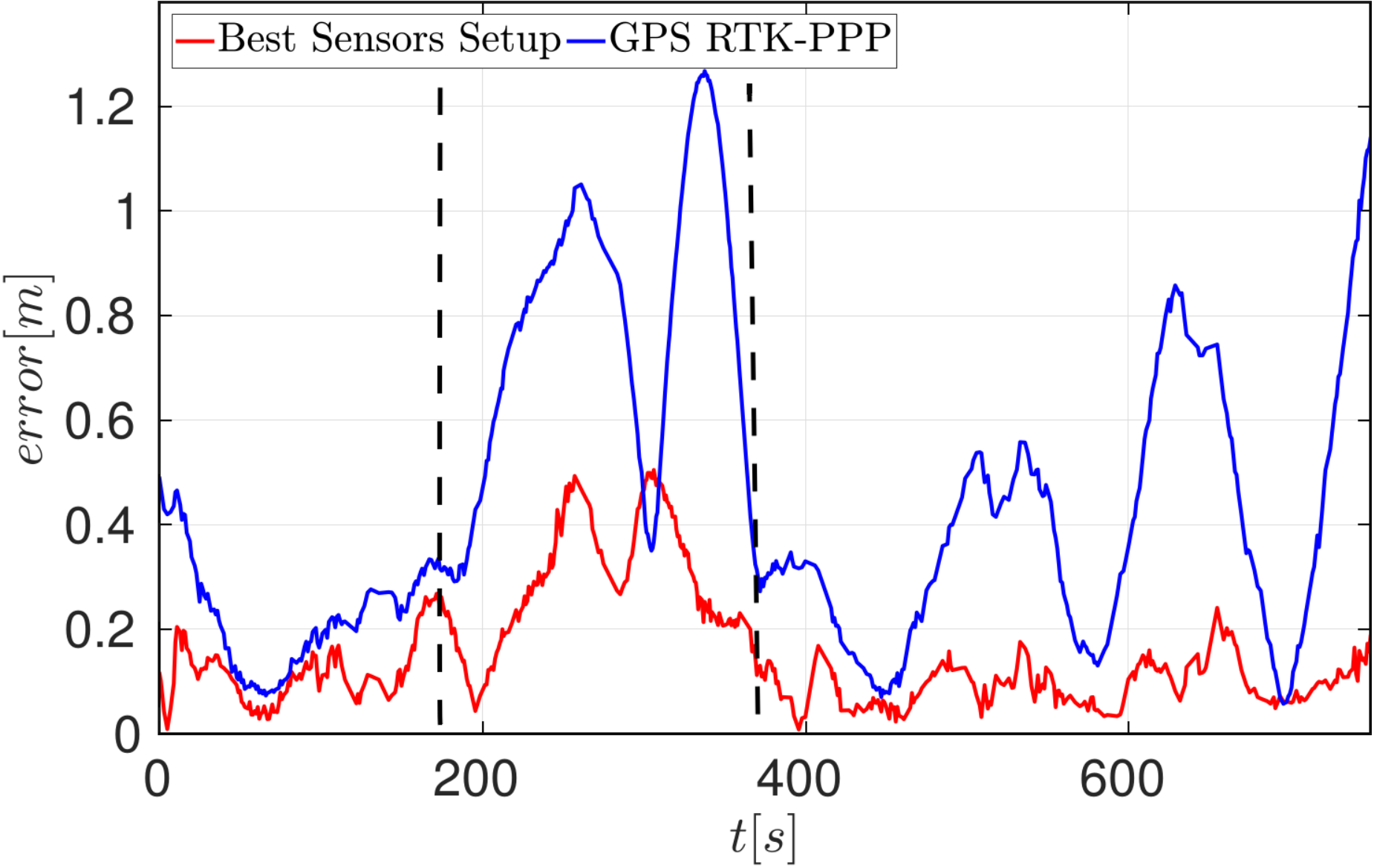}
    \label{fig:datasetC_err}
  \end{subfigure} 
  \caption{(Left) Dataset A, RTK-GPS: Absolute error plots for the raw GPS trajectory and the optimized trajectory obtained by using the best sensors configuration (see Tab.~\ref{tab:err_statistic_datasetA}). (Right) Dataset C, absolute error plots for the raw GPS trajectory and the optimized trajectory (see Tab.~\ref{tab:err_statistic_datasetC}). The time interval when the signal loss happens is bounded by the two dashed lines.}
  \label{fig:dat_topview_comparison}
\end{figure}

This set of experiments shows the effectiveness of the proposed method and the benefits introduced by each cue. We report in Tab.~\ref{tab:err_statistic_datasetA} the results obtained by using different sensor combinations and optimization procedures over \textit{Dataset A} and \textit{Dataset B}. The table is split according to the type of GPS sensor used; the sensor setups that bring the overall best results are highlighted in bold. 
We also compared our system with the ORB SLAM 2 system \cite{murORB2}, a best-in-class Visual SLAM system, with its mapping and loop closures back-ends activated. For a fair comparison, we added the GPS information (PPP and RTK) as a global constraint at each key-frame triggered by ORB SLAM 2.\\
A first result is the positive impact of including the new proposed constraints in the optimization: both the ELEV and MRF cues individually integrated lead to noteworthy improvements in the estimation along the $z$ when a noisy GPS is used (PPP-GPS case). Another remarkable result is the decreasing error trend, almost monotonic: the more sensors we introduced in the optimization process, the smaller the resulting $RMSE$ and $Max$ errors are. This behavior occurs in both \textit{Dataset A} and \textit{Dataset B}, and proves how the proposed method properly handles all the available sources of information. Another important outcome is the relative $RMSE$ improvement obtained between the worst and the best set of cues, which is around the 37\% for RTK case, and 76\% for the PPP case; in both these setups our system outperforms the ORB SLAM 2 system. A noteworthy decrease of the error also happens to the $Max$ error statistic, respectively, 40\% and 70\%: this fact brings a considerable benefit to agricultural applications, where spikes in the location error might lead to harming crops. For the best performing sensor setup, we also report the results obtained by using the SW, on-line pose graph optimization procedure (Sec.~\ref{sec:sliding_window}): also in this case the relative improvement is remarkable (32\% and 67\%, respectively), enabling a safer and more accurate real-time UGV navigation. \\
Fig.~\ref{fig:data_topview_comparison} (top) depicts a qualitative top view comparison between the raw PPP-GPS trajectory (top-left) and the trajectory (top-right) obtained after the pose graph optimization, using the best sensors configuration in \textit{Dataset A}. The error plots (bottom) show how the introduction of additional sensors and constraints allows to significantly improve the pose estimation. Similar results for \textit{Dataset A} and RTK-GPS  are reported in Fig.~\ref{fig:dat_topview_comparison} (left).\\
For both GPSs, the maximal error reduction happens when all the available cues are used within the optimization procedure except for the low cost RTK-GPS case, where the ELEV constraint worsens the error along the $z$ axis.
\begin{figure}[t!]
  \centering
  \begin{subfigure}{\columnwidth}
    \centering
    \includegraphics[width=\textwidth, height=1.9cm]{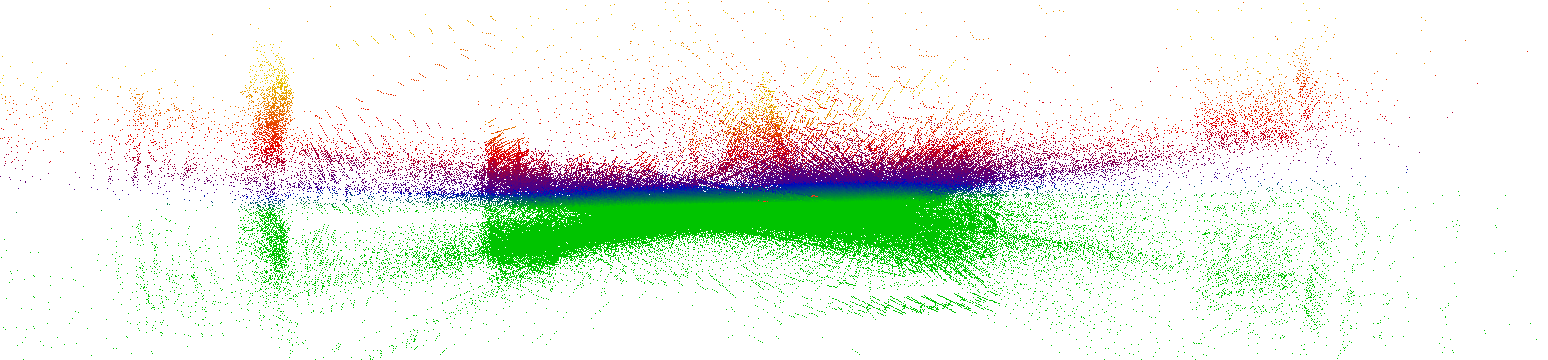}
    \label{fig:with_imu} 
   \end{subfigure}
   \begin{subfigure}{\columnwidth}
    \centering
    \includegraphics[width=.85\textwidth, height=1.9cm]{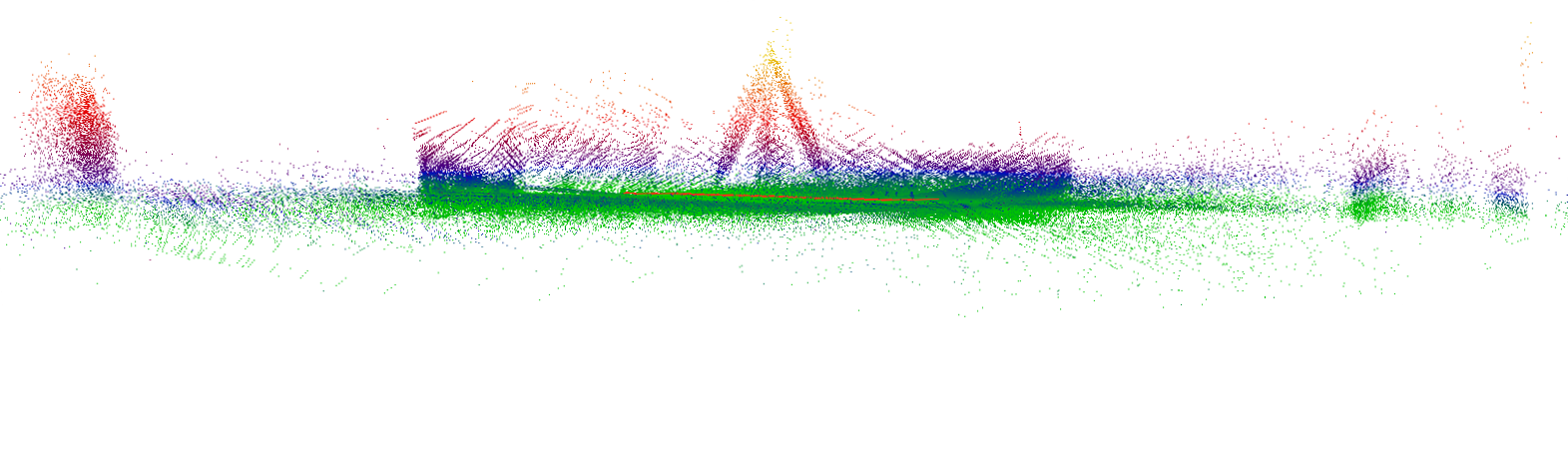}
    \label{fig:without_imu}
  \end{subfigure} 
  \caption{Comparison between output point-clouds: (top) without IMU and LIDAR and (bottom) with IMU and LIDAR in the optimization.}
  \label{fig:clouds}
\end{figure}
Actually, the RTK-GPS usually provides an altitude estimate, which is more accurate than the one provided by the interpolated DEM. It is also noteworthy to highlight the propagation of the improvements among state dimensions: the integration of constraints that only act on a part of the state (e.g., IMU, LIDAR, ELEV) also positively affects the remaining state components.\\
As a further qualitative evaluation, in Fig.~\ref{fig:clouds} we report the global point-cloud obtained by rendering LIDAR scans at each estimated position, with and without the IMU and LIDAR contributions within the optimization procedure: the attitude estimation greatly benefits from these contributions.
The runtimes of our system are reported in Tab.~\ref{tab:global_time}, for both the off-line and on-line, sliding-window cases.
\begin{table}[ht] 
   \scriptsize
   \centering
   \caption{Runtime performance for the global, off-line and the sliding-window (SW), on-line pose graph optimization (Core-i7 2.7 GHz laptop).}
   \label{tab:global_time}
   \begin{tabular}{ cccccc }
  &\footnotesize SW& $\#Nodes$   & $\#Edges$   & $\#Iters$   & $time (s)$ \\\cline{1-6}
  \multirow{2}{*}{{\parbox[c]{1cm}{\centering \textit{Dataset A}}}}
      && 786 & 8259 & 24 & 13.493 \\ 
      &\checkmark& 98 & 763 & 4 & 0.0989 \\ \hline
  \multirow{2}{*}{{\parbox[c]{1cm}{\centering \textit{Dataset B}}}}
      && 754 & 8032 & 22 & 12.993 \\ 
      &\checkmark& 104 & 851 & 5 & 0.1131 \\ \hline
   \end{tabular}
\end{table}

\subsection{Dataset C}

\begin{table}[ht!]
	\scriptsize
	\centering
	\caption{Error statistics in \textit{Dataset C} by using different sensor setups and constraints in the optimization procedure.}
	\label{tab:err_statistic_datasetC}
	  \begin{tabular}{ cccccccccc }
	  \multicolumn{8}{c}{} & \multicolumn{2}{c}{ \textbf{DatasetC} }\\ \cmidrule(lr){9-10} 
	  \rotatebox[origin=c]{90}{\parbox[c]{1cm}{\centering GPS}}&\rotatebox[origin=c]{90}{\parbox[c]{1cm}{\centering WO}} &\rotatebox[origin=c]{90}{\parbox[c]{1cm}{\centering VO}} &\rotatebox[origin=c]{90}{\parbox[c]{1cm}{\centering IMU}} 
	  &\rotatebox[origin=c]{90}{\parbox[c]{1cm}{\centering AMM}} &\rotatebox[origin=c]{90}{\parbox[c]{1cm}{\centering ELEV}}&\rotatebox[origin=c]{90}{\parbox[c]{1cm}{\centering LIDAR}}
	  &\rotatebox[origin=c]{90}{\parbox[c]{1cm}{\centering MRF}} &$Max$ &$RMSE$ \\\cline{1-10}\\
	  \multirow{10}{*}{\rotatebox[origin=c]{90}{\parbox[c]{1cm}{\centering RTK+PPP}}}
	  &&&&&&& &1.313 &0.647\\
	  &\checkmark&&&&&& &1.291 &0.613\\
	  &\checkmark&\checkmark&&&&& &1.259 &0.552\\
	  &\checkmark&\checkmark&&\checkmark&&\checkmark& &1.171 &0.431\\
	  &\checkmark&\checkmark&\checkmark&&&\checkmark& &0.882 &0.356\\
	  &\checkmark&\checkmark&&\checkmark&\checkmark&\checkmark& &0.551 &0.223\\
	  &\checkmark&\checkmark&\checkmark&&&&\checkmark &0.655 &0.204\\
	  &\checkmark&\checkmark&\checkmark&\checkmark&&&\checkmark &0.521 &0.201\\
	  &\checkmark&\checkmark&\checkmark&\checkmark&\checkmark&\checkmark&\checkmark &0.534 &0.181\\
	  &\pmb{\checkmark}&\pmb{\checkmark}&\pmb{\checkmark}&\pmb{\checkmark}&&\pmb{\checkmark}&\pmb{\checkmark} &\textbf{0.419} &\textbf{0.168}\\ \cline{1-10} \noalign{\vskip 1mm}
	  
	  \end{tabular}
\end{table}
This set of experiments is designed to prove the robustness of the proposed system against sudden losses in the GPS sensor accuracy.
In Tab.~\ref{tab:err_statistic_datasetC} we report the quantitative results of our system over \textit{Dataset C} by means of $RMSE$ and $Max$ errors. Even in the presence of a RTK-GPS signal loss that lasts for more than one crop row, the best sensors setup leads to a remarkable $RMSE$ of 0.166 m and a relative improvement around the 72\%. Moreover, also in \textit{Dataset C} the $RMSE$ and the $Max$ error statistics follow the same decreasing trend shown in Tab.~\ref{tab:err_statistic_datasetA}. In Fig.~\ref{fig:dat_topview_comparison} (right) we compare the absolute error trajectories for the best sensors configuration against the error trajectory obtained by using only the GPS measurements: the part where the signal loss occurs is affected by a higher error.
Another interesting observation regards the non-constant effects related to the use of the ELEV constraint. As shown in Tab.~\ref{tab:err_statistic_datasetC}, in some cases it allows to decrease the overall error, while in other cases it worsens the estimate. The latter happens when the pose estimation is reliable enough, i.e. when most of the available constraints are already in use. As explained in section \ref{sec:datA_and_datB}, in such cases the ELEV constraint does not provide any additional information to the optimization procedure, while with a less accurate PPP-GPS its use is certainly desirable. 

\section{Conclusions}
In this paper, we present an effective global pose estimation system for agricultural applications that leverages in a reliable and efficient way an ensemble of cues. We take advantage from the specificity of the scenario by introducing new constraints exploited inside a pose graph realization that aims to enhance the strengths of each integrated information. We report a comprehensive set of experiments that support our claims: the provided localization accuracy is remarkable, the accuracy improvement well scale with the number of integrated cues, the proposed system is able to work effectively with different types of GPS, even in presence of signal degradations. 
The open-source implementation of our system along with the acquired datasets are made publicly available with this paper.


%

\section*{Acknowledgment}

We are grateful to Wolfram Burgard for providing us with the Bosch BoniRob, and to Raghav Khanna and Frank Liebisch to help us in acquiring the datasets.

\ifCLASSOPTIONcaptionsoff
  \newpage
\fi



%


%
%

\bibliographystyle{IEEEtran}
\bibliography{ral_iros_2018_flourish_mapping}




\end{document}